\pdfoutput=1

\documentclass[11pt]{article}

\usepackage{acl}

\usepackage{times}
\usepackage{latexsym}
\usepackage{tabularx}
\usepackage{xcolor}
\usepackage{soul}
\usepackage{amsmath}
\usepackage{amssymb}
\usepackage{booktabs}
\usepackage{tikz}
\usepackage{pgfplots}
\usepackage{caption}
\usepackage{subcaption}

\pgfplotsset{compat=1.17} 
\usepackage[T1]{fontenc}

\usepackage[utf8]{inputenc}

\usepackage{microtype}

\newcommand{\muc}{$\textit{MUC}$}
\newcommand{\bcub}{$\textit{B}^3$}
\newcommand{\ceaf}{$\textit{CEAF}_{e}$}

\newcommand{\emldisplay}[2]{\texttt{\href{mailto:#1}{#2}}}
\newcommand{\eml}[1]{\emldisplay{#1}{#1}}

\newcommand{\mstart}{\texttt{<m>}}
\newcommand{\mend}{\texttt{</m>}}
\newcommand{\tstart}{\texttt{<target>}}
\newcommand{\tend}{\texttt{</target>}}
\newcommand{\estart}{\texttt{<e>}}
\newcommand{\eend}{\texttt{</e>}}
\newcommand{\cstart}{\texttt{<context>}}
\newcommand{\cend}{\texttt{</context>}}

\definecolor{husl0}{rgb}{0.968,0.441,0.536}
\definecolor{husl1}{rgb}{0.809,0.563,0.195}
\definecolor{husl2}{rgb}{0.592,0.642,0.194}
\definecolor{husl3}{rgb}{0.198,0.696,0.400}
\definecolor{husl4}{rgb}{0.210,0.677,0.643}
\definecolor{husl5}{rgb}{0.223,0.657,0.817}
\definecolor{husl6}{rgb}{0.642,0.550,0.958}
\definecolor{husl7}{rgb}{0.960,0.381,0.868}

\definecolor{purple}{rgb}{0.725, 0.36, 0.956}

\title{Efficient Seq2seq Coreference Resolution Using Entity Representations}

\author{Matt Grenander \quad Shay B. Cohen \quad Mark Steedman \\
        School of Informatics \\
        University of Edinburgh \\
        \eml{matt.grenander@ed.ac.uk} \\
        \eml{scohen@inf.ed.ac.uk}, \eml{m.steedman@ed.ac.uk} \\        
}

\begin{document}
\long\def\/*#1*/{}
\maketitle

\begin{abstract}
Seq2seq coreference models have introduced a new paradigm for coreference resolution by learning to generate text corresponding to coreference labels, without requiring task-specific parameters.
While these models achieve new state-of-the-art performance, they do so at the cost of flexibility and efficiency.
In particular, they do not efficiently handle \textit{incremental} settings such as dialogue, where text must processed sequentially.
We propose a compressed representation in order to improve the efficiency of these methods in incremental settings.
Our method works by extracting and re-organizing entity-level tokens, and discarding the majority of other input tokens. 
On OntoNotes, our best model achieves just 0.6 CoNLL F1 points below a full-prefix, incremental baseline while achieving a compression ratio of 1.8.
On LitBank, where singleton mentions are annotated, it passes state-of-the-art performance.
Our results indicate that discarding a wide portion of tokens in seq2seq resolvers is a feasible strategy for incremental coreference resolution.
\end{abstract}

\section{Introduction}
Coreference resolution is a task in which systems identify and resolve expressions referring to the same, real-world entity.
Recently, seq2seq approaches to coreference resolution have reached state-of-the-art abilities on coreference benchmarks and have put forward an exciting new direction for the task \cite{bohnet-etal-2023-coreference, zhang-etal-2023-seq2seq}.
These models leverage encoder-decoder models such as T5 \cite{t5} to directly output a sequence of tokens corresponding to coreference predictions.
Unlike the dominant ``End-to-End'' paradigm \cite{lee-etal-2017-end}, they do not require additional task-specific parameters, and visibly surpass LLM-based approaches \cite{le2024are}.

In this work, we are interested in an \textit{incremental} setting, where the model receives the text piece-by-piece and must detect mentions and compute coreference relations in the given chunk of text before receiving the next one.
This setting more closely imitates human language processing, which is strongly incremental \cite{altmann, christiansen-now-or-never}.
Real-world settings such as dialogue are inherently incremental and would benefit from advances in this domain.

In an incremental setting, existing seq2seq approaches are largely unsatisfying:
\citet{bohnet-etal-2023-coreference} requires the entire output to be re-input for each new sentence in the discourse, while \citet{zhang-etal-2023-seq2seq} assumes the entire document is static and always available.
These limitations motivate us to explore a seq2seq solution that can efficiently perform coreference in an incremental setting.

We hypothesize that re-inputting the entire output at each iteration may be unnecessary for incremental coreference resolution, and that condensing the output text may suffice. 
Using \citet{zhang-etal-2023-seq2seq}'s seq2seq model as a starting point, we propose an \textbf{Entity-Centric representation}.
For each new text chunk, we only re-input text spans corresponding to predicted entities, and discard tokens outside of these spans.
Entities are then re-ordered such that newly mentioned entities appear at the rightmost position of the entity list.

Our model represents a fundamentally different approach to prevailing language modelling trends: we aggressively discard tokens from the context window, and instead keep a ``memory'' representing the current state of the document.

\begin{figure*}[t]
\small
\centering
\begin{tabular}{p{0.95\linewidth}}
\toprule
\textbf{Full-Prefix Incremental:}\\
\textcolor{blue}{<m><m> Wetland Park | 0 </m> workers | 1 </m> are now in the middle of intensive work.<m> They | 1 </m> will complete <m> the park's | 0 </m> entire construction by the beginning of <m> 2006 | 2 </m>, to be able to participate in <m> the <m> 2006 | 2 </m> Discover Hong Kong Year campaign | 3 </m>. We have <m> established | 4 </m> <m> the year 2006 | 2 </m> as Discover Hong Kong Year. Why is <m> that | 4 </m>? Because, as everyone knows, <m> our Disneyland | 5 </m> will open in September of <m> this year | 2 </m>. In addition, we will have Ngong Ping 360, that is, the cable car, er, to the Giant Buddha.} \\
<target> They add to what we already have, like the Avenue of Stars, which is also very famous. Moreover, er, we are including our software. Hong Kong's software is very well known. Like what's used in our Symphony of Lights. We hope to use, er, a variety of hardware and software to package this entire 2006 Discover Hong Kong Year. Without planning it in advance, they chose to settle here. A dream that has been anticipated for more than twenty years will soon come true here.</target> \\
\midrule
\textbf{Entity-Centric Representation:}\\
\textcolor{orange}{<e><m> Wetland Park workers </m> <m> They </m> | 1</e>} \\
\textcolor{orange}{<e><m> Wetland Park</m> <m> the park's </m> | 0 </e>} \\
\textcolor{orange}{<e><m> the 2006 Discover Hong Kong Year campaign </m> | 3 </e>} \\
\textcolor{orange}{<e><m> established </m> <m> that </m> | 4 </e>} \\ \textcolor{orange}{<e><m> our Disneyland </m> | 5 </e>} \\
\textcolor{orange}{<e><m> 2006 </m> <m> 2006</m><m> the year 2006 </m><m> this year </m> | 2 </e>} \\ 
\textcolor{blue}{<context> Because, as everyone knows,<m> our Disneyland | 5 </m> will open in September of <m> this year | 2 </m>. In addition, we will have Ngong Ping 360, that is, the cable car, er, to the Giant Buddha. </context>} \\
<target> They add to what we already have, like the Avenue of Stars, which is also very famous. Moreover, er, we are including our software. Hong Kong's software is very well known. Like what's used in our Symphony of Lights. We hope to use, er, a variety of hardware and software to package this entire 2006 Discover Hong Kong Year. Without planning it in advance, they chose to settle here. A dream that has been anticipated for more than twenty years will soon come true here.</target> \\
\midrule
\textbf{Label:} \\
They add to what we already have, like the Avenue of Stars, which is also very famous. Moreover, er, we are including <m> our software | 6 </m>.<m><m> Hong Kong's | 7 </m> software | 6 </m> is very well known. Like what's used in our Symphony of Lights. We hope to use, er, a variety of hardware and software to package <m> this entire 2006 Discover Hong Kong Year | 3 </m>. Without planning <m> it | 8 </m> in advance, they <m> chose | 8 </m> to settle here. A dream that has been anticipated for more than twenty years will soon come true here. \\
\bottomrule
\end{tabular}
\caption{\textbf{Top}: An example of the Full-Prefix Incremental setting. The text in blue represents previously labelled chunks. The model will predict coreference clusters in the next chunk, enclosed by \tstart\ and \tend\ tokens. 
\textbf{Middle}: An example from the Entity-Centric representation. Mentions from the same cluster are grouped together and labelled with their cluster identity.
The fixed-length context, shown in blue, consists of previously labelled sentences as in the incremental setting.
\textbf{Bottom}: The expected output for this sample.}
\label{fig:example}
\end{figure*}

On OntoNotes \cite{ontonotes}, we find that our method achieves just 0.7 CoNLL F1 below a full-prefix, incremental baseline based on \citet{zhang-etal-2023-seq2seq} while reaching almost twofold compression in the input length.
On LitBank, our model's performance surpasses contemporary methods, while slightly lagging behind the full-prefix baseline by 0.8 F1.
In our analysis, we find that the incremental setting affects the model's ability to predict long-range coreference for named entities and definite noun phrases on OntoNotes.
We hypothesize that incremental processing is particularly difficult on OntoNotes as the dataset does not annotate singletons, resulting in an unfavorable source of noise. 
This finding is reinforced by our results on LitBank, which includes singleton annotation and where we do not see this effect.

\section{Related Work}
Our method builds on recent seq2seq methods for coreference resolution \cite{bohnet-etal-2023-coreference,zhang-etal-2023-seq2seq}.
In these methods, an encoder-decoder model learns to directly map input text to a labelled text representing coreference predictions.

\citet{zhang-etal-2023-seq2seq}'s method explores several annotation schemas for coreference resolution.
Each schema considers a set of added special tokens, corresponding to decisions such as marking mention boundaries and making a clustering prediction.
They fine-tune models from the T5 family \cite{t5} to encode the input document and generate an annotated document corresponding to cluster predictions.
Notably, their method assumes the entire document is always available and they do not explore incremental settings.
We will describe their method in more detail in Section \ref{sec:seq2seq-method}.

Similarly, \citet{bohnet-etal-2023-coreference} fine-tune mT5 \cite{xue-etal-2021-mt5} to output text corresponding to coreference predictions. 
Unlike \citeauthor{zhang-etal-2023-seq2seq}, \citeauthor{bohnet-etal-2023-coreference} process text sentence-by-sentence in a transition-based approach.
The set of actions includes steps such as appending to an existing entity, linking directly to a previous text span, or creating a new entity.
After each action, the system re-inputs the entire output again, meaning computation costs rise considerably for long documents.


Entity representations have appeared in various, non-seq2seq forms in prior work. 
One prominent approach involves maintaining a set of hidden representations corresponding to each entity cluster \cite{yu-etal-2020-cluster, toshniwal-etal-2020-learning, xia-etal-2020-incremental, xu-choi-2022-online, grenander-etal-2022-sentence}, with similar approaches in cross-document coreference resolution \cite{allaway-etal-2021-sequential, logan-iv-etal-2021-benchmarking}.
In contrast to newer seq2seq approaches, they are generally less accurate and more complex, requiring additional task-specific parameters alongside the base encoder and a separate mention detection step (\citeauthor{yu-etal-2020-cluster, toshniwal-etal-2020-learning, xia-etal-2020-incremental, xu-choi-2022-online, grenander-etal-2022-sentence}), or gold mentions as input (\citeauthor{allaway-etal-2021-sequential, logan-iv-etal-2021-benchmarking}).

Our method of compressing and simplifying processed inputs is motivated by cognitive theories of human language processing such as File-Change Semantics \cite{heim-file-change}.
\citeauthor{heim-file-change} argues that discourse processing can be regarded as ``file-keeping'', in which file cards track entities and their properties in each new discourse segment.
Our approach can be seen as implementing a simplified version of this theory, without predicates.

\citet{christiansen-now-or-never} propose that language is rapidly compressed and re-coded as it is encountered.
They argue that the speed humans understand language suggests it is compressed and re-organized, or otherwise forgotten.
Although their mechanism differs from ours, we apply this compression principle in our approach.

Many methods for efficient coreference resolution exist; however, most focus on improving the dominant ``End-to-End'' approach \cite{lee-etal-2017-end}.
For example, \citet{dobrovolskii-2021-word}, \citet{martinelli-etal-2024-maverick} and \citet{kirstain-etal-2021-coreference} improve the efficiency of span representations to ease span comparisons, which is not applicable to seq2seq approaches.
\citet{ahmed-etal-2023-2} focus on efficiency in event coreference resolution but their method largely does not apply to entity resolution.

Outside of coreference, \citet{nawrot2024dynamic} look at retrofitting LLMs to dynamically pool and compress tokens.
Although their motivation is similar to ours, their technique focuses on decoder-only architectures in language modeling. 

\section{Method}
In coreference resolution, a model is provided with a text and predicts a set of entity clusters.
Each cluster denotes a set of co-referring mentions, and each mention marks a unique text span in the document.

We first describe \citet{zhang-etal-2023-seq2seq}'s seq2seq approach (\ref{sec:seq2seq-method}), which serves as the basis for our proposal.
We then describe how we incrementalize their model to provide a comparable baseline in Section \ref{sec:incremental-method}.
Finally, in \ref{sec:entity-method} we explain how we simplify and compress the input using the Entity-Centric representation.

\subsection{Seq2seq Coreference Resolution}\label{sec:seq2seq-method}
\citeauthor{zhang-etal-2023-seq2seq} frames coreference as a text-to-text task, which takes the document as input and outputs the same document annotated with mention boundaries and coreference relations.
Here, we describe their \textbf{Token Action} formulation, referring the reader to \citeauthor{zhang-etal-2023-seq2seq} for other settings.

Token Action augments the vocabulary with special tokens marking mention boundaries and cluster predictions.
Given a mention $(x_i\ \dots\ x_j)$ referring to entity cluster $l_1$, the model outputs:
\[
\mstart\ x_i\ \dots\ x_j\ |\ l_1\ \mend
\]
where $\mstart$ and $\mend$ are special tokens marking the mention start and end, and $|$ is a special token preceding cluster predictions.
The cluster prediction is always an integer greater or equal to 0.

At inference time, the model's output probabilities are modified to 
either output the next token from the document or special tokens.
The outputs are constrained to always form valid coreference predictions, e.g. $|$ must precede $l_1$.


\subsection{Full-Prefix Incremental Baseline}\label{sec:incremental-method}
In \citeauthor{zhang-etal-2023-seq2seq}'s seq2seq formulation, the entire document is encoded in a single pass, with the assumption that the entire document is statically available.
Here, we adapt a full-prefix, incremental baseline where the model receives a text chunk and must compute potential mentions and coreference relations before receiving the next text chunk.

We assume the model has been divided into text chunks $[c_1,\dots,c_N]$, and we denote $c_n$'s labelled sequence as $\hat{c}_n$.
Given text chunk $c_n$, we concatenate all previously labelled text chunks, then append $c_n$ as the target chunk.
We introduce special $\tstart$ and $\tend$ tokens, which indicate the beginning and end of the target chunk.
After processing text chunk $c_n$ and generating predictions $\hat{c}_n$, the input for the next chunk $c_{n+1}$ is
\[
\hat{c}_1\dots\ \hat{c}_n\ \tstart\ c_{n+1} \tend
\]
At test time, we replace each labelled chunk $\hat{c}_i$ with the model's own predictions. An example of the incremental setting is shown in Figure \ref{fig:example}.

Note that \citeauthor{bohnet-etal-2023-coreference} predict coreference relations incrementally in a similar setup.
The main difference lies in their transition actions, which allow the model to predict cluster identities, directly link mentions, or create a singleton mention.

\subsection{Entity-Centric Representation}\label{sec:entity-method}
After processing each text chunk, the Entity-Centric representation compresses the model outputs, keeping tokens from mention spans and discarding all others.
Each entity is represented by its mentions' tokens, including mention bound tokens $\mstart$ and $\mend$.
It is then demarcated with special $\estart$ and $\eend$ tokens, e.g., an entity with mentions $[m_1,\dots,m_k]$ with cluster ID $l_1$ will be:
\[
\estart \mstart\ m_1\ \mend\ \dots\ \mstart\ m_k \mend |\ l_1\ \eend\
\]
New mentions are either appended to their referent cluster or initialized as a new cluster.

We would also like the representation to reflect the notion that recently mentioned entities tend to be more salient in the discourse \cite{grosz-etal-1995-centering}.
After an entity is mentioned, we promote it to the rightmost position of the representation.
This re-ordering signals to the model which entities are likely to be relevant to the current discourse.

We hypothesize that in some cases, solely tracking entities may not provide sufficient context to resolve new mentions.
We experiment with adding a limited context window of previously labelled sentences immediately preceding the target chunk, marked with special $\cstart$ and $\cend$ tokens.
Sentences beyond the window are dropped.

To summarize, suppose after chunks $c_1,\dots,c_n$, the model has observed entity clusters $[1 \dots K]$. Then the input representation for target $c_{n+1}$ is:
\begin{align*}
    & \estart \dots\ |\ l_1\ \eend\ \dots\ \estart \dots\ |\ l_K\ \eend \\
    & \cstart\ \tilde{c}_n\ \cend \\
    & \tstart\ c_{n+1}\ \tend \\
\end{align*}
where $\tilde{c}_n$ consists of the annotated sentences immediately preceding $c_{n+1}$ up to some fixed window length. An example is shown in Figure \ref{fig:example}.

\section{Experiments}\label{sec:experiments}

\subsection{Datasets}\label{sec:datasets}
We train and evaluate our proposed method using the OntoNotes dataset \cite{pradhan-etal-2012-conll}.
The dataset contains annotated coreference chains across 2,802, 343, and 348 documents for training, validation and test splits respectively.
Annotations cover 7 genres such as broadcast news, telephone conversations and weblogs. 

We also train and evaluate on the LitBank dataset \cite{bamman-etal-2020-annotated}, covering literary text extracts. 
The dataset contains 100 documents, each containing over 2000 tokens on average.
Unlike OntoNotes, LitBank includes singleton annotation.
We perform the suggested 10-fold cross validation with an 80-10-10 train-validation-test split, due to the dataset's small size.

\subsection{Metrics}
For evaluation, we follow the standard evaluation in the CoNLL-2012 Shared Task \cite{pradhan-etal-2012-conll} and report \muc\ \cite{vilain-etal-1995-model}, \bcub\  \cite{bagga1998algorithms} and \ceaf\ \cite{luo-2005-coreference} metrics and their average (the CoNLL score).
Each metric measures a distinct aspect of coreference resolution: \muc\ is \textit{link}-based, \bcub\ is \textit{mention}-based and \ceaf\ is \textit{entity}-based.

\subsection{Comparisons}
For all experiments, we use the T0 model \cite{sanh2022multitask}, a 3B parameter encoder-decoder model based on T5 \cite{t5}.

\paragraph{Non-Incremental} As a first baseline and for reproducibility, we re-train \citeauthor{zhang-etal-2023-seq2seq}'s non-incremental Token Action model using T0.
This experiment allows us to more directly compare non-incremental and incremental settings.

\paragraph{Full-Prefix Incremental Baseline} 
We incrementalize the Token Action model using the method described in Section \ref{sec:incremental-method}.
We set the chunk size to 100 tokens, rounded up to the nearest sentence.

\begin{table*}[!t]
\small
\centering
\begin{tabular}{l|c c c c c c c c c|c}
\toprule
&		\multicolumn{3}{c}{\muc}					&	\multicolumn{3}{c}{\bcub}					&	\multicolumn{3}{c}{\ceaf}					&	Avg.	\\
Model	&	P	&	R	&	F1	&	P	&	R	&	F1	&	P	&	R	&	F1	&	F1	\\
\midrule
\citet{le2024are}	&	73.9	&	73.5	&	73.7	&	60.8	&	64.7	&	62.7	&	49.3	&	55.7	&	52.3	&	62.9	\\
\midrule
\citet{joshi-etal-2020-spanbert}	&	85.8	&	84.8	&	85.3	&	78.3	&	77.9	&	78.1	&	76.4	&	74.2	&	75.3	&	79.6	\\
\citet{wu-etal-2020-corefqa}	&	88.6	&	87.4	&	88.0	&	82.4	&	82.0	&	82.2	&	79.9	&	78.3	&	79.1	&	83.1	\\
\citet{kirstain-etal-2021-coreference}	&	86.5	&	85.1	&	85.8	&	80.3	&	77.9	&	79.1	&	76.8	&	75.4	&	76.1	&	80.3	\\
\citet{dobrovolskii-2021-word}	&	84.9	&	87.9	&	86.3	&	77.4	&	82.6	&	79.9	&	76.1	&	77.1	&	76.6	&	81.0	\\
\midrule
\citet{liu-etal-2022-autoregressive}	&	86.1	&	88.4	&	87.2	&	80.2	&	83.2	&	81.7	&	78.9	&	78.3	&	78.6	&	82.5	\\
\citet{bohnet-etal-2023-coreference}, Link-Append	&	87.4	&	88.3	&	87.8	&	81.8	&	83.4	&	82.6	&	79.1	&	79.9	&	79.5	&	83.3	\\
\citet{zhang-etal-2023-seq2seq}, Copy + T0\textsubscript{pp}	&	86.1	&	89.2	&	87.6	&	80.6	&	84.3	&	82.4	&	78.9	&	80.1	&	79.5	&	83.2	\\
\citet{zhang-etal-2023-seq2seq}, Token Action	&	85.9	&	88.6	&	87.2	&	79.6	&	83.5	&	81.5	&	78.9	&	78.0	&	78.5	&	82.4	\\
\midrule
Token Action, Non-Incremental	&	86.1	&	87.9	&	87.0	&	79.8	&	82.2	&	81.0	&	79.1	&	77.3	&	78.2	&	82.0	\\
Token Action, Full-Prefix Incremental	&	86.7	&	84.3	&	85.5	&	80.5	&	77.5	&	79.0	&	78.9	&	70.1	&	74.3	&	79.6	\\
Entity-Centric, Context 0	&	86.7	&	82.7	&	84.6	&	79.8	&	74.8	&	77.3	&	78.7	&	66.9	&	72.3	&	78.1	\\
Entity-Centric, Context 50	&	86.5	&	83.3	&	84.8	&	79.8	&	76.2	&	78.0	&	80.0	&	67.6	&	73.3	&	78.7	\\
Entity-Centric, Context 100	&	87.0	&	83.4	&	85.1	&	80.7	&	75.8	&	78.2	&	79.6	&	68.1	&	73.4	&	78.9	\\
Entity-Centric, Context 200	&	86.8	&	83.3	&	85.0	&	80.4	&	76.2	&	78.2	&	80.0	&	67.9	&	73.5	&	78.9	\\
\bottomrule
\end{tabular}
\caption{Results on the OntoNotes test set. The bottom section shows our proposed methods.}
\label{tab:results}
\end{table*}

\paragraph{Entity-Centric}
We compress the inputs using the Entity-Centric representation described in Section \ref{sec:entity-method}.
We again set the chunk size to 100.
We experiment varying the context length across $\{ 0, 50, 100, 200 \}$.
Using 0 context tokens examines an extreme case where only mention spans and cluster identities are available to the model.

\paragraph{Other Baselines}
We include \citeauthor{zhang-etal-2023-seq2seq}'s highest performing Copy method which uses the 11B parameter T0\textsubscript{pp} model \cite{sanh2022multitask}, as well as \citeauthor{bohnet-etal-2023-coreference}'s Link-Append model, which uses the 13B parameter mT5 model \cite{xue-etal-2021-mt5}.
We also compare against several contemporary non-seq2seq baselines, which typically extend \citet{lee-etal-2017-end}'s E2E approach.
Lastly, we include an LLM-based approach \cite{le2024are}.

\subsubsection{Training}
For all experiments, we use the same hyperparameters as in \citeauthor{zhang-etal-2023-seq2seq}'s Token Action model.
We train each model for 30 epochs using the Hugging Face Transformers library \cite{wolf2019huggingfaces} on 4 NVIDIA RTX A6000 48 GB GPUs.
Each training run takes \string~24 hours.
Note that \citeauthor{zhang-etal-2023-seq2seq} train for 100 epochs, and we do not expect our reproduced baseline to fully recover their scores.

\section{Results}
Table \ref{tab:results} shows our main results on the OntoNotes test set.
First, we note that our reproduction of \citeauthor{zhang-etal-2023-seq2seq}'s Token Action scores 0.4 CoNLL F1 below them.
As mentioned, the difference may be attributed to training epochs, as \citeauthor{zhang-etal-2023-seq2seq} train their models for 100 epochs.
while we train for 30 epochs due to computational limitations.

The \textbf{Full-Prefix baseline} scores 2.4 F1 points lower than the non-incremental counterpart. 
The model sees very little degradation in precision scores, improving on the baseline by 0.6 and 0.7 on \muc\ and \bcub, and dropping 0.2 on \ceaf.
The decrease is mostly in recall performance, where the gap is 3.6, 4.7 and 7.2 for \muc, \bcub\ and \ceaf\ respectively.
\ceaf\ is an entity-focused metric, and the large recall gap suggests that the Full-Prefix baseline tends to miss entire entities relative to the baseline, rather than individual mentions.

The drop in recall scores is unsurprising due to the challenging incremental setting.
This baseline receives chunks of document at a time, and once a decision on anaphoricity has been made, it cannot be revisited after receiving the next chunk.
In contrast, \citeauthor{zhang-etal-2023-seq2seq}'s non-incremental model observes the entire document simultaneously, and can decide more easily if distant mentions co-refer.

The \textbf{Entity-Centric representation} slightly underperforms compared to the Full-Prefix baseline, but overall achieves very close scores.
We find 100 context tokens performs best, achieving only 0.65 CoNLL F1 below the Full-Prefix baseline.
Compared to the Full-Prefix baseline, improvements in precision are slightly offset by declines in recall performance across all three metrics.
The biggest drop is from \ceaf\ recall, where the Entity-Centric model scores 2 points lower.

Using 200 context tokens provides no further gains compared to using 100 context tokens.
This result suggest that some context is important for the Entity-Centric representation, but its usefulness saturates after a point.
On the other hand, using no context at all achieves the lowest score. Impressively, it only suffers a 1.5 CoNLL drop relative to the Full-Prefix baseline, despite discarding all tokens outside of mention spans.

Lastly, we note that \citeauthor{bohnet-etal-2023-coreference}'s Link-Append and \citeauthor{zhang-etal-2023-seq2seq}'s Copy + T0\textsubscript{pp} model outperform all settings. 
Their underlying encoder-decoder models are significantly larger than our own:
\citeauthor{bohnet-etal-2023-coreference} uses the 13B mT5 model \cite{xue-etal-2021-mt5}, while \citeauthor{zhang-etal-2023-seq2seq}'s uses the 11B T0\textsubscript{pp} model \cite{sanh2022multitask}.
Our experiments with the T0 model uses 3B parameters with a similar amount of pre-training data as other T5 models.

\begin{table}[t]
\small
\centering
\addtolength{\tabcolsep}{-0.05em}
\begin{tabular}{l|c c c|c}
\toprule
Model	&	\muc	&	\bcub	&	\ceaf	&	Avg.	\\
\midrule									
\citet{joshi-etal-2020-spanbert}	&	89.5	&	78.2	&	67.6	&	78.4	\\
\citet{toshniwal-etal-2021-generalization}	&	-	&	-	&	-	&	79.3	\\
\citet{zhang-etal-2023-seq2seq}	&	-	&	-	&	-	&	78.3	\\
\midrule									
Non-Incremental	&	88.8	&	77.5	&	68.3	&	78.2	\\
Full-Prefix Incremental 	&	90.3	&	80.3	&	71.6	&	80.7	\\
Entity-Centric, C=100	&	89.9	&	79.0	&	70.9	&	79.9	\\
\bottomrule
\end{tabular}
\caption{Results on the LitBank test set. \citet{joshi-etal-2020-spanbert} is reported by \citet{thirukovalluru-etal-2021-scaling}.}
\label{tab:results-litbank}
\end{table}

\begin{table}[t]
\small
\centering
\addtolength{\tabcolsep}{-0.1em}
\begin{tabular}{l|c|c}
\toprule
Model	&	Peak Mem.	&	Peak Mem. w/o	\\
	&	(GB)	&	Fixed Costs (GB)	\\
\midrule					
Non-Incremental	&	18.1	&	7.4	\\
Full-Prefix Inc.	&	16.2	&	5.5	\\
Entity-Centric, C=100	&	14.5	&	3.8	\\
\bottomrule
\end{tabular}
\caption{GPU memory usage on the LitBank validation set, fold 0. The `Peak Mem. w/o Fixed Costs' column subtracts the memory cost of the model itself, which is constant across the three settings.}
\label{tab:gpu-savings}
\end{table}

Our results on LitBank are shown in Table \ref{tab:results-litbank}. 
The Full-Prefix baseline and Entity-Centric model achieve the highest scores among available systems, with the Entity-Centric model scores 1.6 CoNLL F1 higher than \citet{zhang-etal-2023-seq2seq}.
It drops slightly ($-0.8$ F1) in performance relative to the Full-Prefix baseline, as on OntoNotes.
Interestingly, the performance drop in OntoNotes between incremental and non-incremental systems is not repeated in LitBank, which we explore further in Section \ref{sec:inc-noninc-error-sources}.

\section{Analysis}\label{sec:analysis}
\subsection{Compression Ratio}
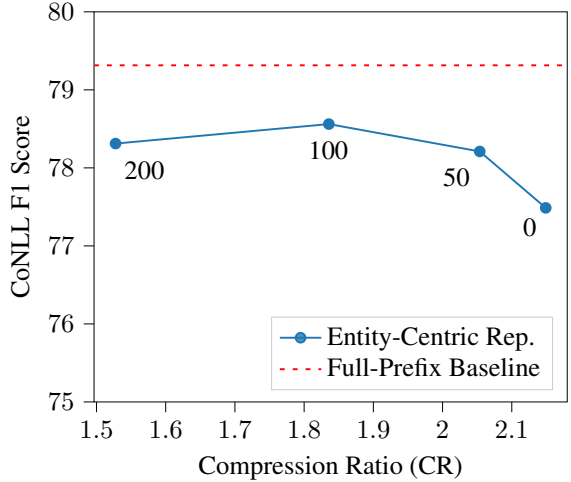
\begin{figure}[t]
    \centering
    \resizebox{\columnwidth}{!}{
    \begin{tikzpicture}

\definecolor{darkgray176}{RGB}{176,176,176}
\definecolor{lightgray204}{RGB}{204,204,204}
\definecolor{steelblue31119180}{RGB}{31,119,180}

\begin{axis}[
legend cell align={left},
legend style={
  fill opacity=0.8,
  draw opacity=1,
  text opacity=1,
  at={(0.97,0.03)},
  anchor=south east,
  draw=lightgray204
},
tick align=outside,
tick pos=left,
x grid style={darkgray176},
xlabel={Compression Ratio (CR)},
xmin=1.49648268725668, xmax=2.18013287644996,
xtick style={color=black},
y grid style={darkgray176},
ylabel={CoNLL F1 Score},
ymin=75, ymax=80,
ytick style={color=black}
]

\addplot [thick, steelblue31119180, mark=*]
table {%
1.52755769585638 78.3114255160774
1.83572021288772 78.5613206103874
2.05372420029931 78.2104597061591
2.14905786785026 77.4872213526725
};
\addlegendentry{Entity-Centric Rep.}
\addplot [thick, red, dash pattern=on 2pt off 3.3pt]
table {%
1.49648268725668 79.3144255513743
2.18013287644996 79.3144255513743
};
\addlegendentry{Full-Prefix Baseline}

\node [below right] at (axis cs:  1.527, 78.21) {200};
\node [below] at (axis cs:  1.835, 78.46) {100};
\node [below left] at (axis cs: 2.053, 78.11) {50};
\node [below left] at (axis cs: 2.149, 77.48) {0};

\end{axis}

\end{tikzpicture}    
    }
    \caption{Compression Ratio experiments on the OntoNotes validation set.}
    \label{fig:compression}
\end{figure}

Since the Entity-Centric model only keeps a fraction of the total input at each iteration, its input length grows much slower as new chunks are processed.
Here, we quantify the reduction in size of the Entity-Centric model's inputs relative to the Full-Prefix baseline.
We define the compression ratio (CR) as the ratio of the Full-Prefix baseline's input length to Entity-Centric's input length on the last target chunk of each document.
We compute the ratio for each document in the OntoNotes validation set, and report the average in Figure \ref{fig:compression}.

The highest performing Entity-Centric model, using 100 context tokens, achieves CR=1.8 while scoring 0.75 F1 points below the Full-Prefix baseline.
On the right side of the figure, higher compression is achieved at the expense of performance: when no context tokens are used, the Entity-Centric representation achieves CR=2.1 while retaining 97.7\% of Full-Prefix baseline performance.

\subsection{GPU Memory Usage}
More practically, we can also ask whether the Entity-Centric model reduces GPU memory usage compared to other baselines.
Since the Entity-Centric model requires keeping less tokens in context, we expect GPU memory usage should decrease.
To test this hypothesis, we use the \verb|torch.cuda.memory._record_memory_|
\verb|history| function in PyTorch to record GPU memory usage while running inference, then run the Non-Incremental, Full-Prefix Incremental and Entity-Centric (C=100) models on the LitBank, fold 0 validation set. 
We inspect the memory usage\footnote{https://docs.pytorch.org/memory\_viz} to determine maximum memory usage as well as the memory footprint of the models themselves (i.e. without processing documents).

Table \ref{tab:gpu-savings} shows the GPU usage of the three settings.
On the right column, we only consider the cost of document processing, i.e. without the fixed cost of the model itself, which is constant (10.7 GB) across the three settings.
Comparing the Non-Incremental and Entity-Centric models, maximum GPU usage decreases from 7.4 to 3.8 GB, representing a 1.9x reduction, similar to our compression ratio analysis.
The result shows that the Entity-Centric is a practical method for reducing GPU memory usage.

\subsection{Entity Ordering}
\begin{table}[t]
\centering
\small
\begin{tabular}{l|c c c c}
\toprule
Setting	&	\muc	&	\bcub	&	\ceaf	&	Avg.	\\
\midrule									
Entity-Centric 	&	84.3	&	77.9	&	78.6	&	78.6	\\
Document Order	&	84.3	&	77.7	&	78.4	&	78.4	\\
\bottomrule
\end{tabular}
\vspace{-0.1cm}
\caption{Results on entity ordering on the OntoNotes validation set. Entity-Centric refers to the regular Entity-Centric representation, while Document Order drops the entity re-ordering step. Columns denote F1 scores.}
\label{tab:doc-order}
\end{table}

The Entity-Centric model dynamically re-orders entities each time a mention is detected.
We experiment with ablating this step: we train the model as usual with 100 context tokens, but instead of re-ordering, we keep entities in the same order as they appear in the document.
The result is shown in Table \ref{tab:doc-order}.

We find that entity re-ordering has a small effect, lifting the CoNLL F1 score by 0.18. 
We also evaluate the `Document Order' ablation on the test set and find a larger drop of 0.45 CoNLL F1 relative to the Entity-Centric model, indicating it plays a small effect in coreference performance.

\subsection{Sources of Error in Incremental vs. Non-Incremental Settings}\label{sec:inc-noninc-error-sources}
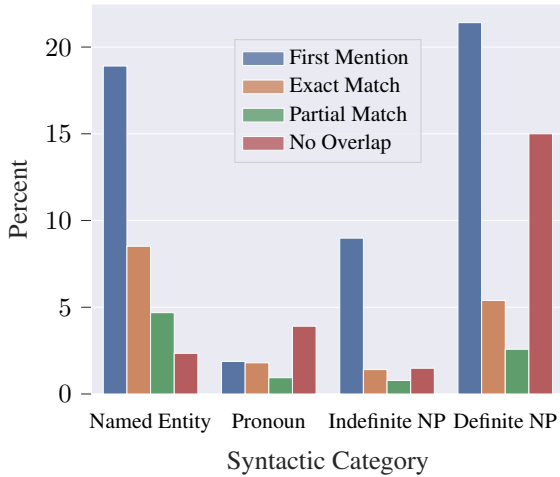
\begin{figure}[t]
\centering
\resizebox{\columnwidth}{!}{
\begin{tikzpicture}

\definecolor{darkslategray38}{RGB}{38,38,38}
\definecolor{indianred1819295}{RGB}{181,92,95}
\definecolor{lavender234234242}{RGB}{234,234,242}
\definecolor{lightgray204}{RGB}{204,204,204}
\definecolor{mediumseagreen95157109}{RGB}{95,157,109}
\definecolor{peru20313699}{RGB}{203,136,99}
\definecolor{steelblue88116163}{RGB}{88,116,163}

\begin{axis}[
axis background/.style={fill=lavender234234242},
axis line style={white},
legend cell align={left},
legend style={
  font=\small,
  fill opacity=0.8,
  draw opacity=1,
  text opacity=1,
  at={(0.5,0.91)},
  anchor=north,
  draw=lightgray204,
  fill=lavender234234242
},
tick align=outside,
x grid style={white},
xlabel=\textcolor{darkslategray38}{Syntactic Category},
xmajorticks=true,
xmin=-0.5, xmax=3.5,
xtick style={draw=none},
xtick={0,1,2,3},
xticklabels={Named Entity,Pronoun,Indefinite NP,Definite NP},
x tick label style={font=\small},
y grid style={white},
ylabel=\textcolor{darkslategray38}{Percent},
ymajorgrids,
ymajorticks=true,
ytick pos=left,
ymin=0, ymax=22.4765625,
ytick style={color=darkslategray38}
]
\draw[draw=white,fill=steelblue88116163] (axis cs:-0.4,0) rectangle (axis cs:-0.2,18.90625);
\draw[draw=white,fill=steelblue88116163] (axis cs:0.6,0) rectangle (axis cs:0.8,1.875);
\draw[draw=white,fill=steelblue88116163] (axis cs:1.6,0) rectangle (axis cs:1.8,8.984375);
\draw[draw=white,fill=steelblue88116163] (axis cs:2.6,0) rectangle (axis cs:2.8,21.40625);
\draw[draw=white,fill=peru20313699] (axis cs:-0.2,0) rectangle (axis cs:2.77555756156289e-17,8.515625);
\draw[draw=white,fill=peru20313699] (axis cs:0.8,0) rectangle (axis cs:1,1.796875);
\draw[draw=white,fill=peru20313699] (axis cs:1.8,0) rectangle (axis cs:2,1.40625);
\draw[draw=white,fill=peru20313699] (axis cs:2.8,0) rectangle (axis cs:3,5.390625);
\draw[draw=white,fill=mediumseagreen95157109] (axis cs:-2.77555756156289e-17,0) rectangle (axis cs:0.2,4.6875);
\draw[draw=white,fill=mediumseagreen95157109] (axis cs:1,0) rectangle (axis cs:1.2,0.9375);
\draw[draw=white,fill=mediumseagreen95157109] (axis cs:2,0) rectangle (axis cs:2.2,0.78125);
\draw[draw=white,fill=mediumseagreen95157109] (axis cs:3,0) rectangle (axis cs:3.2,2.578125);
\draw[draw=white,fill=indianred1819295] (axis cs:0.2,0) rectangle (axis cs:0.4,2.34375);
\draw[draw=white,fill=indianred1819295] (axis cs:1.2,0) rectangle (axis cs:1.4,3.90625);
\draw[draw=white,fill=indianred1819295] (axis cs:2.2,0) rectangle (axis cs:2.4,1.484375);
\draw[draw=white,fill=indianred1819295] (axis cs:3.2,0) rectangle (axis cs:3.4,15);
\draw[draw=white,fill=steelblue88116163] (axis cs:0,0) rectangle (axis cs:0,0);
\addlegendimage{area legend, draw=white,fill=steelblue88116163}
\addlegendentry{First Mention}

\draw[draw=white,fill=peru20313699] (axis cs:0,0) rectangle (axis cs:0,0);
\addlegendimage{area legend,draw=white,fill=peru20313699}
\addlegendentry{Exact Match}

\draw[draw=white,fill=mediumseagreen95157109] (axis cs:0,0) rectangle (axis cs:0,0);
\addlegendimage{area legend,draw=white,fill=mediumseagreen95157109}
\addlegendentry{Partial Match}

\draw[draw=white,fill=indianred1819295] (axis cs:0,0) rectangle (axis cs:0,0);
\addlegendimage{area legend,draw=white,fill=indianred1819295}
\addlegendentry{No Overlap}

\end{axis}

\end{tikzpicture}    
}
\caption{Mentions correctly predicted by the non-incremental model but missed by the Full-Prefix Incremental model, divided by syntactic category and string overlap with antecedent.}
\label{fig:syn-errors}
\end{figure}
In our OntoNotes results, we find a 2.5 F1 point gap between incremental and non-incremental settings.
While not directly comparable, \citeauthor{bohnet-etal-2023-coreference}'s system does not suffer from any performance gap compared to \citeauthor{zhang-etal-2023-seq2seq}'s non-incremental model, despite also implementing an incremental setting.
This observation leads us to examine the main sources of error in our Full-Prefix baseline.

We collect all mentions in the OntoNotes validation set that are correctly detected by the non-incremental baseline but missed by the Full-Prefix model.
Inspired by \citet{otmazgin-etal-2023-lingmess}, we divide mentions into four broad syntactic categories: (1) named entities, (2) pronouns, (3) indefinite NPs and (4) definite NPs.\footnote{We use the OntoNotes gold POS tags and set of simple heuristic rules to assign each mention to its syntactic category.}
For each mention, we further record whether the mention's direct antecedent is an exact string match, a partial string match\footnote{We record a partial string match if either the mention or its antecedent are substrings of each other.} or otherwise has no string overlap. 
For example, the mention \textit{he} with antecedent \textit{John} will be marked as a pronoun with no overlap to its antecedent, while mention \textit{John} with antecedent \textit{John Doe} will be marked as a named entity with partial overlap to its antecedent.
Since the first mention of an entity has no antecedent, we record these mentions as a separate category.

The results are shown in Figure \ref{fig:syn-errors}.
Named entities and definites dominate the missed mention set, accounting for 76\% of all mentions.
Although certain categories are not easy to remedy, such as definite noun phrases with no overlap to their antecedent, we expect other categories, such as named entities with exact match, to be relatively straightforward.
Assuming their antecedent is also missed, named entities and definite noun phrases with exact or partial matches to their antecedent account for 40\% of missed mentions.

This error type reflects a fundamental difficulty of incremental settings: the model must decide early on in a text whether a given span of text is likely to be involved in coreference and cannot revisit the decision in later chunks.
The OntoNotes annotation schema compounds this difficulty: a mention is only marked if it co-refers with another mention in the text, rather than if it simply \textit{could} be the target of coreference.
These ``singleton'' mentions are occasionally annotated in other coreference datasets \cite{uryupina-etal-2016-arrau,yu-etal-2022-codi, bamman-etal-2020-annotated}, and we note that we do not experience this drop in performance on LitBank, which includes singletons.
This artifact means the model does not learn whether a given span of text \textit{could be the target of coreference} but rather whether it is \textit{likely to have been annotated as such in the current document}.
We suspect this error type is a major cause of lower recall scores, in particular the entity-focused \ceaf\ metric.

Interestingly, \citeauthor{bohnet-etal-2023-coreference} does not seem to suffer similarly, despite also being an incremental model.
We suspect this difference is due to their distinct set of transition actions.
\citeauthor{bohnet-etal-2023-coreference} allows for a `Link' action, which marks two text spans as co-referring without first creating a mention.
However, the technique cannot be extended to our entity-centric representations, as it requires keeping the entire document in context at all times.
 
Lastly, we carry out oracle experiments on adding back gold coreference links to the Full-Prefix model across several steps: (1) named entities that are exact matches of their antecedent, (2) named entities that are partial matches of their antecedent, (3) definite noun phrases that are exact matches of their antecedent, and (4) definite noun phrases that are partial matches of their antecedent.

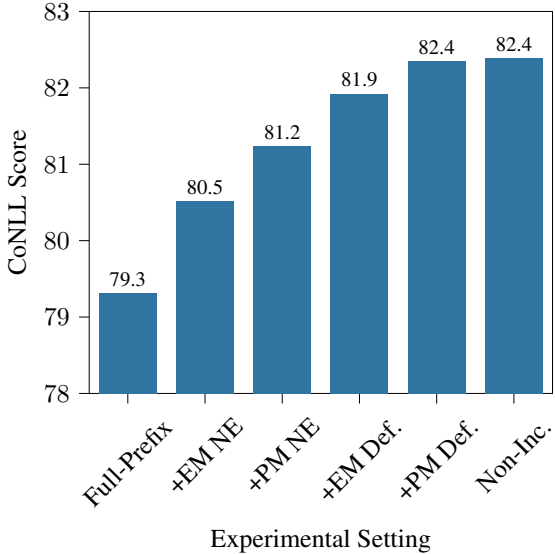
\begin{figure}[t]
\centering
    \resizebox{\columnwidth}{!}{
\begin{tikzpicture}

\definecolor{darkgray176}{RGB}{176,176,176}
\definecolor{darkslategray66}{RGB}{66,66,66}
\definecolor{steelblue49115161}{RGB}{49,115,161}

\begin{axis}[
tick align=outside,
tick pos=left,
unbounded coords=jump,
x grid style={darkgray176},
xlabel={Experimental Setting},
xmin=-0.5, xmax=5.5,
xtick style={color=black},
xtick={0,1,2,3,4,5},
xticklabels={Full-Prefix,+EM NE,+PM NE,+EM Def.,+PM Def.,Non-Inc.},
x tick label style={rotate=45},
y grid style={darkgray176},
ylabel={CoNLL Score},
ymin=78, ymax=83,
ytick style={color=black}
]
\draw[draw=none,fill=steelblue49115161] (axis cs:-0.375,0) rectangle (axis cs:0.375,79.31);
\draw[draw=none,fill=steelblue49115161] (axis cs:0.625,0) rectangle (axis cs:1.375,80.51);
\draw[draw=none,fill=steelblue49115161] (axis cs:1.625,0) rectangle (axis cs:2.375,81.23);
\draw[draw=none,fill=steelblue49115161] (axis cs:2.625,0) rectangle (axis cs:3.375,81.92);
\draw[draw=none,fill=steelblue49115161] (axis cs:3.625,0) rectangle (axis cs:4.375,82.35);
\draw[draw=none,fill=steelblue49115161] (axis cs:4.625,0) rectangle (axis cs:5.375,82.39);

\draw (axis cs:0,79.31) ++(0pt,0pt) node[
  scale=0.8,
  anchor=south,
  text=black,
  rotate=0.0
]{79.3};
\draw (axis cs:1,80.51) ++(0pt,0pt) node[
  scale=0.8,
  anchor=south,
  text=black,
  rotate=0.0
]{80.5};
\draw (axis cs:2,81.23) ++(0pt,0pt) node[
  scale=0.8,
  anchor=south,
  text=black,
  rotate=0.0
]{81.2};
\draw (axis cs:3,81.92) ++(0pt,0pt) node[
  scale=0.8,
  anchor=south,
  text=black,
  rotate=0.0
]{81.9};
\draw (axis cs:4,82.35) ++(0pt,0pt) node[
  scale=0.8,
  anchor=south,
  text=black,
  rotate=0.0
]{82.4};
\draw (axis cs:5,82.39) ++(0pt,0pt) node[
  scale=0.8,
  anchor=south,
  text=black,
  rotate=0.0
]{82.4};

\end{axis}

\end{tikzpicture}    
    }
    \caption{Results on the OntoNotes validation set after adding gold coreference links corresponding to specified syntactic categories.
    Non-Inc.=Non-Incremental, EM=Exact Match, PM=Partial Match, NE=Named Entities, Def.=Definite noun phrases.}
    \label{fig:oracle-add-ner}
\end{figure}

The results are shown in Figure \ref{fig:oracle-add-ner}.
At each step, the CoNLL F1 increases, with the largest increase (1.2 F1) coming from adding exact match named entities.
After adding back gold mentions across all four steps, the resulting CoNLL score equals the non-incremental baseline's performance.

\subsection{NER-Augmented Inference}\label{sec:ner-augmented}
Since our results show that simply adding named entities with an exact string match to their antecedent can significantly improve performance, we carry out two further oracle experiments on adding NER information to the Full-Prefix model during inference.
In both approaches, we augment the Full-Prefix model's predicted clusters with information from the gold NER labels in OntoNotes.\footnote{We provide additional information on the types of NER labels in Appendix \ref{sec:add-ner-results-full}.}

Our first method involves forcibly marking NER spans at inference time.
During inference, if the current token is the left bound of a mention in the NER annotation layer, we make the model generate a mention start token $\mstart$. 
We then allow the model to continue generating as normal, including the mention end token $\mend$ and the cluster ID.\footnote{We also experimented with forcing the mention end token but found it did not lead to better results.}

The second approach takes a more conservative approach to augmenting predicted mentions.
After running inference as usual, we add all mentions in the NER annotation layer that are either (1) exact string matches of each other or (2) an exact string match with a mention in an existing cluster.

The results on the OntoNotes validation set are shown in Table \ref{tab:add-ner-results}.
We find that although F1 scores increase, the improvements are very slight.
In the full results,\footnote{The full recall and precision scores can be found in Appendix \ref{sec:add-ner-results-full}.} we find that although recall scores increase, precision scores across all three metrics decrease.
The decrease in precision is particularly surprising, since we expect the method to only add high-quality exact match named entities.

We analyze errors made by the second approach on 20 documents in the OntoNotes validation set. 
We find that in 40\% of cases, the mention in the NER layer occurs \textit{within} the coreference layer mention; e.g. if the coreference layer contains post-modifiers such as ``\textit{Nelson Mandela}'' vs. ``\textit{Nelson Mandela, who was an anti-apartheid activist}''.

In another 40\% of cases, we find the annotators miss marking two exact match named entities.
This annotation error is particularly noticeable when the named entities occur far from each other.

The remaining 20\% of errors occur due to inadvertently including named entities in copular relations, e.g. ``My mother was Thelma Wahl''.
The OntoNotes annotation schema only includes the left side of copular relations, meaning named entities on the right side should not be included.

The success of prior models in predicting the noise in long-distance named entity coreference annotations suggests, to some degree, that they have learned a model of idiosyncrasies in OntoNotes annotations rather than to superior performance as a model of coreference itself.

\begin{table}[t]
\centering
\small
\begin{tabular}{l|c c c}
\toprule
Model	&	Prec.	&	Rec.	&	F1	\\
\midrule							
Non-Incremental Baseline	&	82.4	&	82.4	&	82.4	\\
\midrule							
Full-Prefix Incremental	&	82.0	&	76.8	&	79.3	\\
 + NER Forced Mention Start	&	80.5	&	78.6	&	79.6	\\
 + NER Exact String Match	&	81.4	&	78.1	&	79.7	\\
 + Pseudosingletons, 30K	&	81.7	&	78.2	&	79.9	\\
 + Pseudosingletons, 60K	&	81.0	&	80.0	&	80.5	\\
\bottomrule
\end{tabular}
\caption{Scores after augmenting inference with NER labels in the Full-Prefix setting. Columns denote F1.}
\label{tab:add-ner-results}
\end{table}

\subsection{Training with Pseudosingletons}
Another potential solution to address OntoNotes' lack of singleton annotation is to augment the training dataset with predicted singleton mentions.
We experiment with \citet{toshniwal-etal-2021-generalization}'s approach with labeling \textit{pseudosingletons} using a trained mention detector.
\citet{toshniwal-etal-2021-generalization} use the \textit{longdoc} model based on the non-incremental SpanBERT \cite{joshi-etal-2020-spanbert} encoder to score all text spans in each document in OntoNotes, based on which are likely to be mentions.
Pseudosingletons are then selected from the top-scoring spans outside of gold mentions.
Since the true number of singletons is unclear, they experiment with adding 30K, 60K or 120K pseudosingletons, and publicly release the annotations.
Using their labels, we experiment with adding 30K and 60K pseudosingletons.
In each experiment, we augment the OntoNotes training dataset with pseudosingletons, then train the Full-Prefix Incremental model as normal on the augmented dataset.

The results are shown in Table \ref{tab:add-ner-results}.
Similar to the NER-augmented inference approaches in Section \ref{sec:ner-augmented}, we find that adding pseudosingletons results in a small increase in F1, and gains in recall are offset by losses in precision.
Our best-performing setting, with 60K pseudosingletons, results in a 3.2\% increase in recall but a decrease in precision of 1.0\%.
Overall, the pseudosingleton augmentation method improves over NER-augmented methods, but still underperforms compared to the non-incremental baseline.

As in the NER Exact String Match method, we analyze mention span errors in the 60K pseudosingleton experiment, finding many similar error cases.
The pseudosingletons occasionally include slightly mismatched mention spans, for example, \textit{Panama}, when the correct mention span is \textit{Panama's}.
As before, annotation error also affects the method's effectiveness. For example, annotators correctly label a named entity (\textit{Rodrigo Miranda}) but fail to mark the pronouns \textit{he} and \textit{him} as co-referring.
The pseudosingleton method then incorrectly labels these pronouns as singletons, leading to unwanted noise during training.

\section{Conclusion}
We present a compressed representation for incremental coreference resolution.
Our method stores and organizes tokens corresponding to entities in the text, while discarding other tokens outside of a small context window.
On OntoNotes, the Entity-Centric model scores 0.6 F1 below the Full-Prefix Incremental baseline's performance, while compressing the input by a ratio of 1.8.
On LitBank, the Entity-Centric model outperforms other SOTA approaches, demonstrating that compressing input representations can be successfully applied to the literary domain while reaching SOTA performance.

Incremental systems display a marked difference in performance relative to SOTA approaches on OntoNotes vs. LitBank datasets.
In our analysis, we show the gap between non-incremental and incremental models on OntoNotes is dominated by named entities and definites.
We hypothesize these cases are due to the lack of singleton annotation in OntoNotes, and we do not observe this pattern on the singleton-annotated LitBank dataset.
Adding gold coreference links corresponding to exact/partial match named entities and definites improves performance; however, we find augmenting inference with NER and pseudosingletons during training struggles with annotation noise and mismatched spans.

\newpage
\section*{Limitations}
This work both trains and evaluates on the benchmark OntoNotes dataset.
OntoNotes may be considered a good dataset for generalizability due to including 7 distinct genres in its data; however, there is no guarantee models trained on this data will necessarily generalize to out-of-domain examples.
Several works have researched the difficulties of generalization in coreference resolution models \cite{subramanian-roth-2019-improving,xia-van-durme-2021-moving,toshniwal-etal-2021-generalization, yuan-etal-2022-adapting}.
These works have specifically noted generalization issues with models trained on OntoNotes.

One frequent issue, which we also discuss in our work, is the lack of singleton annotation in OntoNotes.
The absence of this annotation means models trained on OntoNotes will usually fail to predict singletons in other datasets where this type of annotation is present.
In our work, we also show this lack of annotation is also harmful beyond simply detecting singleton mentions, causing additional difficulty for detecting mentions in an incremental setting.

\section*{Ethical Considerations}
NLP systems must take special consideration to not demonstrate harmful towards protected groups.
Coreference resolution systems have previously been shown to learn stereotypical associations from training corpora. 
\citet{bolukbasi} show that word embeddings may exhibit gender stereotypes, such as associating ``receptionist'' with ``female''.

Coreference resolution models may also learn gender biases, and methods exist to counter this effect \cite{rudinger-etal-2018-gender, zhao-etal-2018-gender}.
Any deployment of our proposed model should first should first evaluate on specialized datasets such as \citet{webster-etal-2018-mind} to determine whether the system unfairly predicts certain labels.

\section*{Acknowledgements}
The authors acknowledge the use of resources provided by the Isambard-AI National AI Research Resource (AIRR). Isambard-AI is operated by the University of Bristol and is funded by the UK Government’s Department for Science, Innovation and Technology (DSIT) via UK Research and Innovation; and the Science and Technology Facilities Council [ST/AIRR/I-A-I/1023]. \cite{isambard}

\bibliography{anthology,anthology2,custom}

\newpage
\appendix
\section{Qualitative Error Analysis on Entity-Centric Models}
\begin{figure}[t]
\centering
\small
\begin{tabular}{p{0.94\linewidth}}
\toprule
The owner of the nursing home Thelma Wahl was being taken to tells \textcolor{red}{\textbf{\{\ us \}\textsubscript{1}}} by the time the bus arrived \dots \\
\dots \\
\textcolor{red}{\textbf{\{\ CNN\ \}\textsubscript{2}}}\ made repeated attempts to contact administrators \dots \\
\bottomrule
\end{tabular}
\caption{An example error from the Entity-Centric model not made by the Full-Prefix Incremental baseline. The Entity-Centric model only stores the mention ``us'', without additional context, meaning the model cannot infer that ``us'' refers to ``journalists from CNN''. It subsequently marks ``CNN'' as a separate cluster.}
\label{fig:qualitative}
\end{figure}

We conduct a qualitative error analysis on a sample of documents from the OntoNotes validation and characterize errors made by the Entity-Centric model but which are correctly predicted by the Full-Prefix Incremental baseline.
Although in general, the predictions and errors between the two systems are the same, we find the removal of context can occasionally lead to incorrect predictions.
For example, in Figure \ref{fig:qualitative}, the mentions ``us'' and ``CNN'' both refer to journalists in the given piece. 
However, since the context is removed after the mention ``us'' is detected, the model later identifies ``CNN'' as a separate entity, without realizing the story is being told from the perspective of journalists. 
In this case, adding speaker tags to the Entity-Centric representation may offer a simple solution.

\section{Entity Ordering Recall and Precision Scores}
\label{sec:document-reorder-full}
The full precision and recall scores for the entity ordering experiments are shown in Table \ref{tab:document-reorder-full}.

\begin{table*}[t!]
\small
\centering
\begin{tabular}{l|c c c c c c c c c|c}
\toprule
& 		\multicolumn{3}{c}{\muc}					&	\multicolumn{3}{c}{\bcub}					&	\multicolumn{3}{c}{\ceaf}					&	Avg.	\\
Setting	&	Prec.	&	Rec.	&	F1	&	Prec.	&	Rec.	&	F1	&	Prec.	&	Rec.	&	F1	&	F1	\\
\midrule
Entity-Centric Representation	&	86.5	&	82.2	&	84.3	&	80.6	&	75.3	&	77.9	&	79.2	&	68.5	&	78.6	&	78.6	\\
Document Order	&	87.1	&	81.6	&	84.3	&	81.3	&	74.4	&	77.7	&	79.5	&	67.8	&	78.4	&	78.4	\\
\bottomrule
\end{tabular}
\caption{Results on the OntoNotes validation set ablating different entity ordering methods.}
\label{tab:document-reorder-full}
\end{table*}

\section{NER-Augmented Inference Additional Information}
\label{sec:add-ner-results-full}
In our experiments with NER-augmented inference, we explored using several different sets of NER categories.
We found that certain NER types in OntoNotes, such as ordinal and cardinal numbers, rarely or never intersect with mentions in the coreference annotations.
After analyzing the degree of overlap between each NER category and mentions in the coreference annotations, we found using the GPE, PERSON and ORG tags in the first oracle experiment (with Forced Mention Start), and a set of ten categories in the second experiment, resulted in the best scores.
The ten categories are: GPE, ORG, PERSON, LAW, FAC, LANGUAGE, EVENT, PRODUCT, LOC, DATE and WORK\_OF\_ART.

Lastly, the full precision and recall scores for the NER-Augmented Inference experiments are shown in Table \ref{tab:add-ner-results-full}.

\begin{table*}[t!]
\small
\centering
\begin{tabular}{l|c c c c c c c c c|c}
\toprule
&		\multicolumn{3}{c}{\muc}					&	\multicolumn{3}{c}{\bcub}					&	\multicolumn{3}{c}{\ceaf}					&	Avg.	\\
Model	&	P	&	R	&	F1	&	P	&	R	&	F1	&	P	&	R	&	F1	&	F1	\\
\midrule
Non-Incremental	&	86.5	&	87.1	&	86.8	&	81.2	&	82.2	&	81.7	&	79.5	&	77.9	&	78.7	&	82.4	\\
Incremental	&	86.6	&	83.3	&	84.9	&	80.6	&	77.1	&	78.8	&	78.9	&	70.0	&	74.2	&	79.3	\\
 + NER Forced Mention Start	&	85.4	&	84.7	&	85.0	&	79.2	&	78.9	&	79.0	&	77.1	&	72.3	&	74.6	&	79.6	\\
 + NER Exact String Match	&	86.0	&	84.3	&	85.2	&	80.0	&	78.3	&	79.1	&	78.2	&	71.7	&	74.8	&	79.7	\\
 + Pseudosingletons, 30K	&	86.5	&	84.0	&	85.2	&	80.9	&	77.9	&	79.4	&	77.6	&	72.8	&	75.2	&	79.9	\\
 + Pseudosingletons, 60K	&	85.7	&	85.3	&	85.5	&	79.8	&	79.9	&	79.9	&	77.4	&	74.8	&	76.1	&	80.5	\\
\bottomrule
\end{tabular}
\caption{Results on the OntoNotes validation set with different methods on incorporating gold NER labels.}
\label{tab:add-ner-results-full}
\end{table*}

\end{document}